\documentclass[acmsmall]{acmart}
\setcopyright{none}
\renewcommand\footnotetextcopyrightpermission[1]{}

\acmConference{}{}{}
\acmBooktitle{}
\acmYear{}
\acmDOI{}
\acmISBN{}
\acmPrice{}

\usepackage{cite}
\usepackage{amsmath,amsfonts}
\usepackage{graphicx}
\usepackage{textcomp}
\usepackage{xcolor}
\definecolor{custompurple}{rgb}{0.5, 0.2, 0.7} 
\usepackage{algorithm}
\usepackage{microtype}
\usepackage{subfigure}
\usepackage{booktabs} 
\usepackage{hyperref}
\usepackage{algpseudocode}
\usepackage{setspace}
\usepackage{subcaption}
\usepackage{multirow}

\usepackage{mathtools}
\usepackage{amsthm}
\usepackage{pifont}
\usepackage{soul}

\newcommand{\cmark}{\ding{51}} 
\newcommand{\xmark}{\ding{55}}
\AtBeginDocument{%
 \let\authorsaddresses\relax
  }

\makeatletter
\def\@journalName{}
\def\@journalVolume{}
\def\@journalNumber{}
\def\@journalArticle{}
\def\@journalYear{}
\def\@journalMonth{}
\makeatother

\author{Abhishek Kumar Mishra}
\affiliation{%
  \department{Electrical and Computer Engineering}
  \institution{Drexel University}
  \city{Philadelphia}
  \state{PA}
  \country{USA}
}
\email{am4862@drexel.edu}

\author{Arya Somasundaram}
\affiliation{%
  \institution{UCLA}
  \city{Los Angeles}
  \state{California}
  \country{USA}
}
\email{arysom992@gmail.com}

\author{Anup Das}
\affiliation{%
  \department{Electrical and Computer Engineering}
  \institution{Drexel University}
  \city{Philadelphia}
  \state{PA}
  \country{USA}
}
\email{ad3639@drexel.edu}

\author{Nagarajan Kandasamy}
\affiliation{%
  \department{Electrical and Computer Engineering}
  \institution{Drexel University}
  \city{Philadelphia}
  \state{PA}
  \country{USA}
}

\title{Efficient Aspect Term Extraction using Spiking
Neural Network}

\setcopyright{none}
\renewcommand\footnotetextcopyrightpermission[1]{}
\begin{document}
\thispagestyle{empty}
\makeatletter

\def\@ACM@manuscript@footer{}

\def\@ACM@manuscript@text{}

\thispagestyle{plain}
\pagestyle{plain}

\makeatother

\begin{abstract}
Aspect Term Extraction (ATE) identifies aspect terms in review sentences, a key subtask of sentiment analysis. While most existing approaches use energy-intensive deep neural networks (DNNs) for ATE as sequence labeling, this paper proposes a more energy-efficient alternative using Spiking Neural Networks (SNNs). Using sparse activations and event-driven inferences, SNNs capture temporal dependencies between words, making them suitable for ATE. The proposed architecture, SpikeATE, employs ternary spiking neurons and direct spike training fine-tuned with pseudo-gradients. Evaluated on four benchmark SemEval datasets, SpikeATE achieves performance comparable to state-of-the-art DNNs with significantly lower energy consumption. This highlights the use of SNNs as a practical and sustainable choice for ATE tasks.
\end{abstract}




\maketitle

\makeatletter
\def\@ACM@manuscript@footer{}
\makeatother

\section{Introduction}\label{sec:introduction}
\begin{figure}[t!]
  \centering
  \includegraphics[width=0.99\linewidth]{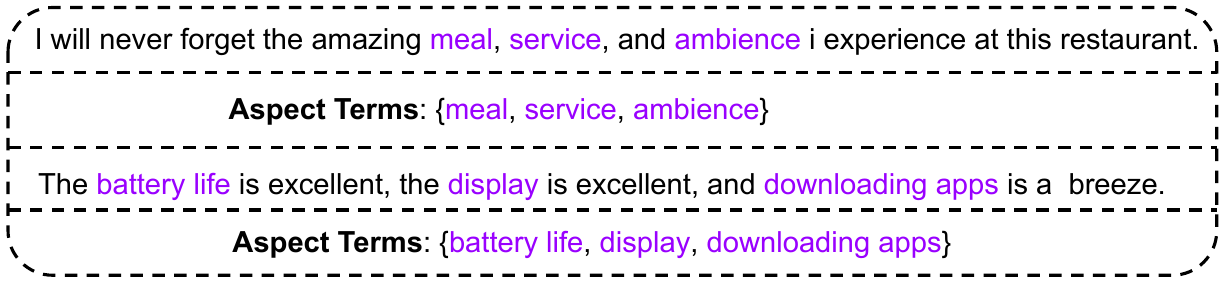}
  \caption{Review sentences illustrate the ATE task where aspect terms are highlighted in purple.}
  \label{fig:ATETask}
  \vspace{-12pt}
\end{figure}

Aspect term extraction (ATE) is an essential subtask of aspect-based sentiment analysis (ABSA) that aims to identify specific aspects terms within a review in which users have expressed their opinions (Fig.~\ref{fig:ATETask}). Detecting and understanding these terms in user-generated content is vital for numerous applications. For example, analyzing customer sentiment from reviews on e-commerce web platforms can offer insights to improve products, brands, and services, and improve marketing strategies for businesses and organizations. Given the vast amount of online content, manually identifying aspect information is impractical. 
Therefore, an automated ATE system is needed to pinpoint opinions associated with specific aspects within unstructured text. This fine-grained analysis is central to ABSA, as it enables accurate representations of
user sentiment and preferences~\citep{peng2020knowing, zhang2023target}.

Over the past two decades, ATE has become a well-studied problem within natural language processing, particularly for recommendation systems and information retrieval. Beyond simply identifying product or domain characteristics, ATE supports related tasks such as aspect co-occurrence analysis, domain profiling, and aspect summarization~\citep{zhao2020spanmlt}. With advances in deep learning, recent approaches treat ATE as a sequence labeling or token classification task, leading to the development of large and complex neural network models designed to capture fine-grained details in text~\citep{liu2015fine, xu2018double, xu2019bert, yang2020constituency, mishra2021does,chen2020enhancing, mao2021joint}.

Deep neural network (DNN) models have proven highly effective in sequence labeling tasks due to their ability to learn directly from raw data and their strong generalization capabilities. However, as DNNs grow in size and complexity to achieve better performance, they require substantial computational resources for training and inference, raising concerns about environmental sustainability. This limitation becomes particularly challenging when DNNs are deployed on devices with limited energy resources, where high energy consumption can be prohibitive~\citep{lv2022spiking,assaraf2022openai, matinizadeh2024fully, matinizadeh2024neuromorphic, mishra2025verification, matinizadeh2024fully1, mishra2024wafer2spike, achiam2023gpt}.

An alternative solution worth exploring is the use of \emph{Spiking Neural Networks} (\emph{SNNs}) which are designed using the operational principles of biological neurons where communication between neurons occurs via short impulses or spikes transmitted through synapses. These spiking neurons can be arranged in feedforward layers or in recurrent topologies~\citep{ghosh2009spiking}. This brain-inspired approach to computing holds considerable promise for spatial and temporal pattern recognition tasks~\citep{lv2022spiking,cao2015spiking, wu2018spatio, mishra2024wafer2spike, mishra2022built, mishra2023online, varshika2023hardware,matinizadeh2024neuromorphic,zhu2023spikegpt}.  SNNs process information through discrete spikes, which are well-suited to handle high-dimensional spatial patterns. This sparse processing approach also reduces noise and enables the network to capture subtle details and variations within patterns. When paired with neuromorphic hardware~\citep{davies2018loihi,akida1,matinizadeh2024neuromorphic,mishra2024model,kumar2025hierarchical}, SNNs can achieve improved accuracy with extreme computational efficiency, making them a compelling option
for sustainable AI solutions.

By developing energy-efficient SNN models for ATE, this work contributes to a more sustainable infrastructure for web-based NLP tasks. ATE, as part of sentiment analysis, is crucial for analyzing web content such as reviews, blogs, and social networks. Low-power SNN models reduce the resource demand for these tasks, which, at scale, lessens the web's environmental footprint and aligns with the growing demand for sustainable technology practices. Therefore, the SNN-based approach for ATE aims to advance Natural Language Processing (NLP) in an energy-conscious direction while enriching the Web by supporting accessible, scalable, and environmentally sustainable web infrastructure.

Although SNNs provide a more sustainable alternative, their adoption and performance in certain tasks still trail behind those of DNNs. \emph{This paper presents the first application of SNNs for ATE task and demonstrates that a well-trained SNN can achieve performance comparable to current DNN-based SOTA solutions.} To address the unique challenges associated with SNNs---extreme sparsity, non-differentiable operators, reliance on approximate gradients, and single-bit
activations---we propose a novel architecture designed to support effective training and convergence. Our approach includes specialized input encoding, innovative network design with two-bit activations, and tailored training strategies that together address these challenges. The contributions of this work are the following.
\begin{itemize}
\item We develop an SNN-based approach for ATE that advances NLP in an energy-conscious direction and enriches the Web by supporting accessible, scalable, and environmentally sustainable infrastructure.

\item We develop SpikeATE, a domain-specific SNN for accurate recognition of aspect terms. The sparse spiking mechanism of the network and simplified calculations result in notable computational efficiency.

\item We develop a convolutional spike encoding layer that transforms token-based embeddings into spike representations while preserving the spatial and temporal dependencies in the original embeddings. Backpropagation is also customized to support training within SNN.

\item We develop a ternary spiking neuron model that outputs values of \(\{-1, 0, 1\}\) instead of the traditional \(\{0, 1\}\), to serve as the fundamental processing unit within the SNN. Doing so significantly increases the neuron’s information capacity. It also improves the model's control over the influence of each type of spike on the membrane potential, preventing excessive activation, and maintaining stability by balancing excitatory and inhibitory signals. 
\end{itemize}
SpikeATE's performance in terms of token-classification accuracy is evaluated on the Semantic Evaluation benchmark dataset~\citep{pontiki-etal-2014-semeval, pontiki-etal-2015-semeval, pontiki-etal-2016-semeval}. The results show that it achieves competitive performance compared to DNNs while achieving significant computational efficiency, as estimated through a widely-used power consumption model. SpikeATE achieves a reduction in power consumption of up to 41x over traditional DNN models. The SpikeATE code repository and the dataset used to obtain these results are available
publicly at \url{https://github.com/abhishekkumarm98/SpikeATE}.

This paper is structured as follows. Section~\ref{sec:related_work} discusses related work on aspect-term extraction. Section~\ref{sec:background} provides an overview of sequence labeling and the foundational concepts of SNNs, including training methods for these networks. Section~\ref{sec:ternary} describes the proposed ternary spiking neuron model, and Section~\ref{sec:arch} develops the SpikeATE architecture. Section~\ref{sec:evaluation} describes the experimental setup
and key results, and Section~\ref{sec:conclusion} concludes the paper.

\section{Related Work}\label{sec:related_work}
Aspect term extraction has been well studied over the past two decades. Early approaches relied on predefined rules, hand-crafted features, or prior knowledge to address ATE~\citep{qiu2011opinion,liu2013opinion}. With advances in deep learning, neural network methods such as LSTM, CNN, and BERT have become dominant, allowing richer feature representations for ATE~\citep{liu2015fine,xu2018double,xu2019bert}. ATE can be approached in both supervised and unsupervised ways. For unsupervised ATE, neural networks incorporate topic modeling~\citep{he2017unsupervised, ouyang2023unsupervised}. Supervised research emphasizes the development of neural sequence taggers~\citep{liu2015fine, xu2018double, chen2020enhancing, wang2021progressive}. A recent trend uses unified frameworks, where ATE is combined with tasks such as extracting opinion terms and describing the sentiment at the aspect level to improve performance~\citep{mao2021joint,li2019unified}.


An ongoing challenge for ATE is the lack of annotated data, especially as the models grow larger and more complex. To address this, Li et al. introduced a conditional data augmentation approach\citep{li2020conditional} while Xu et al. fine-tuned BERT in domain-specific datasets to improve sequence labeling performance~\citep{xu2019bert}. Furthermore, Chen et al. tackled data sparsity by incorporating soft prototypes trained on internal or external data~\citep{chen2020enhancing} and Wang et al. used self-training with unlabeled data to mitigate the problem~\citep{wang2021progressive}.

Compared to DNNs, SNNs can achieve competitive, and sometimes superior, performance for various spatio-temporal classification
tasks~\citep{cao2015spiking, wu2018spatio, lv2022spiking, zhu2023spikegpt}. This success is attributed to the distinctive capabilities of spiking neurons, which dynamically encode information across multiple timescales, activate sparsely, and enable event-driven inference. Although some studies have demonstrated the effectiveness of SNNs in NLP tasks~\citep{diehl2016conversion,lv2022spiking,rao2022long,zhu2023spikegpt}, their application in this domain remains relatively underexplored.

Diehl et al. used pre-trained word embeddings in a TrueNorth implementation of a recurrent neural network (RNN), achieving an accuracy of 74\% for a question-classification task~\citep{diehl2016conversion}. (TrueNorth is a neuromorphic hardware architecture developed by IBM.) Rao et al. demonstrated that long- and short-term memory units (LSTM) can be implemented within spike-based neuromorphic hardware using spike frequency adaptation, allowing their model to perform a question-answer task~\citep{rao2022long}. Lv et al. proposed a two-step method (conversion and fine-tuning) to train spiking convolutional neural networks for sentiment classification, introducing a simple but effective encoding of pre-trained word embeddings as spike trains~\citep{lv2022spiking}. For transformer-based SNNs, Zhu et al. developed SpikeGPT, a spiking language model for both natural language generation and understanding tasks, achieving efficient sentiment analysis and classification with fewer high-cost operations~\citep{zhu2023spikegpt}. 

Although the above approaches address generative and text classification tasks, the potential of SNNs in aspect term extraction has not been explored. To fill this gap, we introduce SpikeATE, a unique SNN architecture tailored for efficient ATE, which addresses the specific challenges SNNs face: extreme sparsity, non-differentiable operators, reliance on approximate gradients, and single-bit activations, which together complicate training convergence. Our approach uses input encoding to represent word embeddings as spike trains, a ternary spiking neuron design, optimized network architecture, and custom training strategies to effectively overcome these challenges.
\section{Background}\label{sec:background}
We describe ATE as a token-level classification task and familiarize the reader with basic concepts related to SNNs. 

\subsection{Token-Level Classification}\label{subsec:background}
Assume a tokenized sentence $S^i=\{tk_1, tk_2, \dots, tk_R\}$ of sequence length $R$. In a token-level classification framework, the ATE model receives $S^i$ as input and outputs a label sequence $Y^i = \{y_1, y_2, \dots, y_R\}$, where $y \in \{B, I, O\}$. The labels $\{B, I, O\}$ denote whether each token is at the beginning, inside or outside of an aspect term, respectively. Given a training set $\mathbf{T_D} = \{(S^i, Y^i)\}_{i=1}^N$, containing $N$ examples, where each $S^i$ is a tokenized review sentence associated with its corresponding label sequence $Y^i$, the goal is to develop an spiking ATE model that can accurately predict these label sequences and effectively perform the ATE task. To predict labels $\{B, I, O\}$, we use a supervised learning approach to train the SpikeATE model. The model is defined as follows:
\begin{equation}
\small
    \begin{aligned}
    SpikeATE(\overrightarrow{S} = [tk_1, tk_2, \dots, tk_l]) \longrightarrow \{ Y \in \{O, B, I\} \mid 
    O = 0, B = 1, I = 2 \}
    \end{aligned}
\end{equation}
This equation represents the SpikeATE model, where the input token sequence $\overrightarrow{S}$ produces the predicted output label sequence $\hat{Y}$. Since the label sequence $Y$ consists of discrete values $\{O, B, I\}$, this task is treated as a classification problem in which the model learns to map token sequences to their corresponding labels.

\subsection{Spiking Neural Network Architecture}\label{subsec:snn}
\begin{figure}[t!]
\centering
\begin{minipage}{0.35\textwidth}{\includegraphics[width=\textwidth]{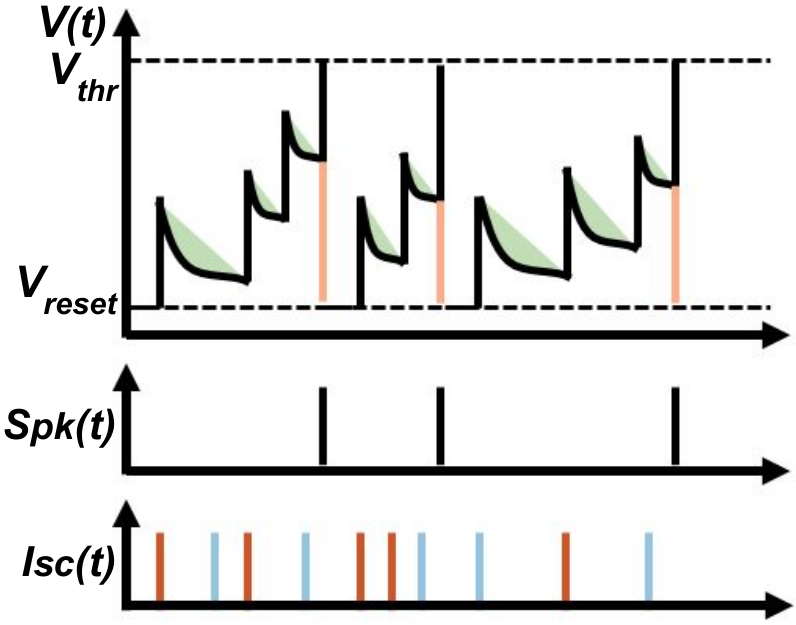}}
\end{minipage}
\begin{minipage}{0.55\textwidth}
    \centering
    \vspace{-3em}  
    \begin{equation} \label{eq:Eq1}
        {{Spk}_t}^{L_j} = 
        \begin{cases} 
        1$ \& ${V_t}^{L_j} = V_{reset}, & \text{if } {V_t}^{L_j} \ge V_{thr}; \\
        0, & \text{if } {V_t}^{L_j} < V_{thr}
        \end{cases} 
    \end{equation}
    \hfill
    \begin{align} \label{eq:Eq2}
        {Isc}^L_t &= W_{scd}^L \cdot {Isc}^L_{t-1} + W_{fv} \cdot {Spk}^{L-1}_t, \; \mathrm{and} \nonumber \\
        {V_t}^L &= W_{vd}^L \cdot V_{t-1}^L + {Isc}^L_t
    \end{align}
\end{minipage}
\caption{The update, fire, and reset cycle for an LIF neuron.}
\label{fig:LIF_Info}
\vspace{-12pt}
\end{figure}

An SNN is represented as a graph, where a vertex represents a spiking neuron, and each directed edge represents a synapse linking these neurons. The weight of the edge represents the strength of the synaptic connection between two neurons. To build an SNN, a range of neuronal models can be used. We have chosen a current-based leaky integrate-and-fire (LIF) model as the computing unit~\citep{izhikevich2004model,ghoreishee2025new,ghoreishee2025improving}. LIF neurons aggregate input signals through a weighted summation, which influences the membrane potential of the $j^{\text{th}}$ neuron, denoted as ${V_t}^{L_j}$, in the $L^{\text{th}}$ layer at time $t$. When this combined input drives the membrane potential above a set threshold, $V_{\text{thr}}$, the neuron emits a spike ${Spk}_t$, and its membrane potential ${V_t}^{L_j}$ is reset to a baseline value, $V_{\text{reset}}$, which in our case is zero as shown in \eqref{eq:Eq1}. Figure~\ref{fig:LIF_Info} shows the behavior of the neuron, where the color-coded spikes indicate the input provided by different presynaptic neurons that are connected to this neuron. 

The membrane potential of a neuron can be modeled using a resistor-capacitor (RC) circuit, which represents the membrane's capacity to charge and discharge in response to electrical impulses. This approach simulates how neurons react to synaptic inputs. By solving the differential equations that govern the RC circuit, we can approximate the dynamics of the membrane potential over time, as shown in \eqref{eq:Eq2}. The equations govern the mechanism by which input spikes are converted into synaptic current, \( {Isc}^L_t \), subsequently affecting the membrane potential, \( {V_t}^L \), in the \( L^{\text{th}} \) layer
at time \( t \). Here, \( {W_{scd}^L} \in \mathbb{R}^+ \) represents the matrix of synaptic current decay factors, applied to each LIF neuron in the \( L^{\text{th}} \) layer. The component $ W_{fv} \cdot {Spk}^{L-1}_t$ produces
the postsynaptic potential defined for the \( L^{\text{th}} \) layer, while \( {Spk}^{L-1}_t \) represents spike events from neurons in the preceding layer and $W_{fv} \in \mathbb{R}$ is a trainable parameter of postsynaptic function. Additionally, \( W_{vd}^L \) denotes the voltage decay factor for each neuron in the \( L^{\text{th}} \) layer. To optimize the performance of the model and minimize the reliance on predefined hyperparameters, both \( W_{scd}^L \) and \( W_{vd}^L \) are designed as trainable parameters within our architecture.

\subsection{Training Spiking Neural Networks}
Training methods fall into two main categories: direct spike training methods and conversion-based DNN-to-SNN methods. In the direct method using supervised learning, SNNs are trained using gradient descent along with backpropagation using spike timing information or specialized forms of backpropagation designed for spikes, such as backpropagation over time~\citep{wu2018spatio, neftci2019surrogate}. Due to the non-differentiability of spiking neurons, surrogate gradients are used to guide backpropagation~\citep{lee2020enabling, neftci2019surrogate}. For unsupervised learning of SNNs, spike-timing-dependent plasticity-based rules have received considerable attention~\citep{diehl2015unsupervised}. 

Conversion-based methods replace the activations that occur in DNNs with neuron firing rates within the equivalent SNN~\citep{cao2015spiking,sengupta2019going}, or by substituting ReLU activations for spiking neurons~\citep{bu2023optimal}. These methods address the challenge posed to direct training methods due to non-differentiability of spikes. DNN-to-SNN conversion is an approximation that places a considerable load on the layer-to-layer bandwidth due to the binary nature of spike activations. This leads to reduced accuracy in the resulting SNNs and requires additional simulation time steps to convert high-precision activations into spikes accurately, thereby increasing inference latency. \emph{Our approach employs a spike-based direct supervised training method with a backpropagation technique specifically optimized for spike-based learning.}


\section{Ternary Spiking Neuron}\label{sec:ternary}
Since SNNs convey information using binary spikes, \{0,1\}, they achieve high energy efficiency by replacing multiplications of activations and weights with simple additions. However, binary spike activation maps (feature maps) have limited information capacity compared to the full-precision activation maps of DNNs. This limitation hinders their ability to effectively carry information from membrane potentials, as they quantize membrane potential distributions across different SNN layers to the same spike values, leading to loss of information and reduced precision~\citep{sun2022deep, guo2022recdis, guo2024ternary}. To address this issue, we develop a model for a ternary spiking neuron that enhances the information capacity by using \{-1,0,1\} spikes to transmit information. It retains the same event-driven, multiplication-free benefits as binary spiking neurons and is represented as
{\small
\begin{align} \label{eq:Ternary}
&\overset{+}{Spk}^{L-1}_t = ({Spk}^{L-1}_t==1)*{Spk}^{L-1}_t \;, \; \; \overset{-}{Spk}^{L-1}_t = ({Spk}^{L-1}_t==-1)*{Spk}^{L-1}_t, \nonumber \\
&{Isc}^L_t = W_{scd}^L \cdot {Isc}^L_{t-1} + \overset{+}W_{fv} \cdot \overset{+}{Spk}^{L-1}_t + \overset{-}W_{fv} \cdot \overset{-}{Spk}^{L-1}_t \;, \; \;   
 {V_t}^L = W_{vd}^L \cdot V_{t-1}^L + {Isc}^L_t \;,\; \;  \mathrm{and}\\ 
&{{Spk}_t}^{L_j} = 
\begin{cases} 
1$ \& ${V_t}^{L_j} = V_{reset},  \text{if } {V_t}^{L_j} \ge V_{thr}; \\
-1$ \& ${V_t}^{L_j} = V_{reset},  \text{if } {V_t}^{L_j} \le -V_{thr}; \\
0, \text{if } -V_{thr} < {V_t}^{L_j} < V_{thr}
\end{cases} \nonumber
\end{align} 
}
Here, $\overset{+}{Spk}^{L-1}_t \in \{0,1\}$ and $\overset{-}{Spk}^{L-1}_t \in \{-1,0\}$ represent positive and negative spiking feature maps extracted from the previous layer, with $\overset{+}W_{fv}$ and $\overset{-}W_{fv}$ as their respective trainable weight parameters for the postsynaptic potential functions. By applying distinct weights for positive and negative spikes, the ternary spiking neuron processes excitatory (positive) and inhibitory (negative) activations separately. This approach improves the model's control over the impact of each type of spike on the membrane potential, preventing excessive activation, and maintaining stability through balanced excitatory and inhibitory signals. It also increases the SNN's representational capacity, allowing it to capture complex features.

In comparison, Sun et al. \citep{sun2022deep} use \(\{0, 1, 2\}\) spikes but require multiplications, whereas our model achieves multiplication-free inference to maintain high computational efficiency while distinguishing between excitatory and inhibitory activations. Unlike Guo et al. \citep{guo2024ternary}, who also use ternary spikes, our approach applies distinct weights for positive and negative spikes, further enhancing stability and representational capacity. This is especially beneficial in the ATE task, as it enables the model to capture nuanced features that reflect both positive and negative sentiments within reviews, thus providing a more refined understanding of aspect-based sentiment. 



\section{Development of SpikeATE}\label{sec:arch}
\begin{figure}[t!]
\centering
\includegraphics[width=0.99\textwidth]{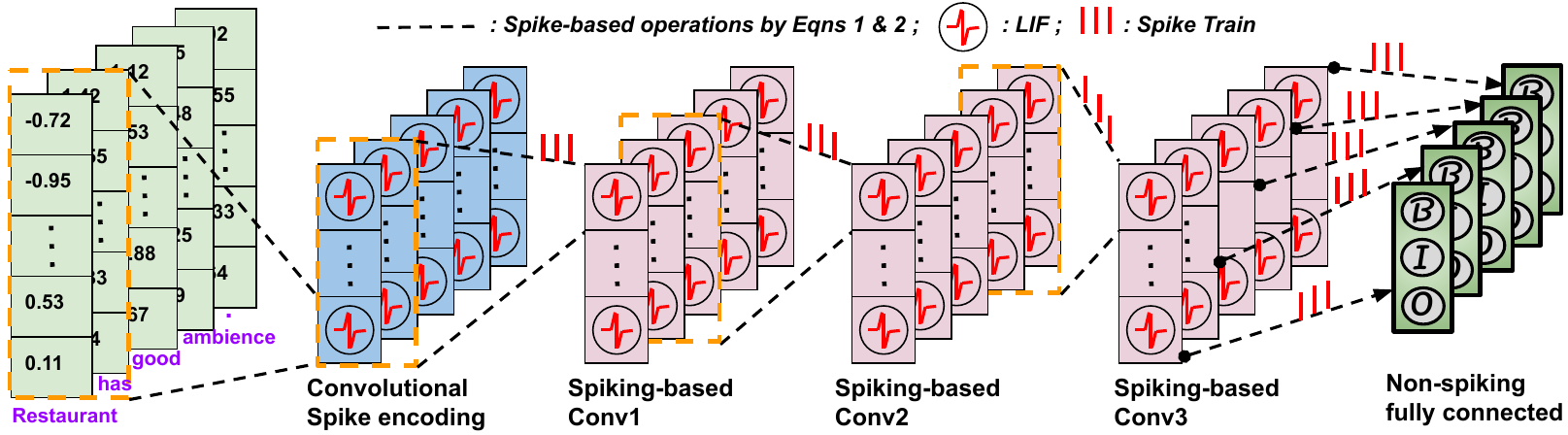}
\caption{The SpikeATE architecture consisting of a convolutional spike encoding layer, multiple spike-based convolutional layers, and the output layer.} 
\label{fig:SpikeATE}
\vspace{-12pt}
\end{figure}

The SpikeATE architecture shown in Fig.~\ref{fig:SpikeATE} consists of a convolutional spike encoding layer to generate spikes from the sequential input data, followed by multiple spike-based convolutional layers to extract spatio-temporal features, terminated by a fully connected output layer. 

\subsection{Convolutional Spike Encoding Layer} 
Given tokenized input data, embeddings are used to generate spikes with the appropriate frequency and interspike intervals. This encoding converts sequential input data into dynamic temporal spike patterns and is inspired by previous studies~\citep{neftci2019surrogate, rathi2020diet}. 

\begin{figure}[t!]
\centering
\includegraphics[width=0.88\textwidth]{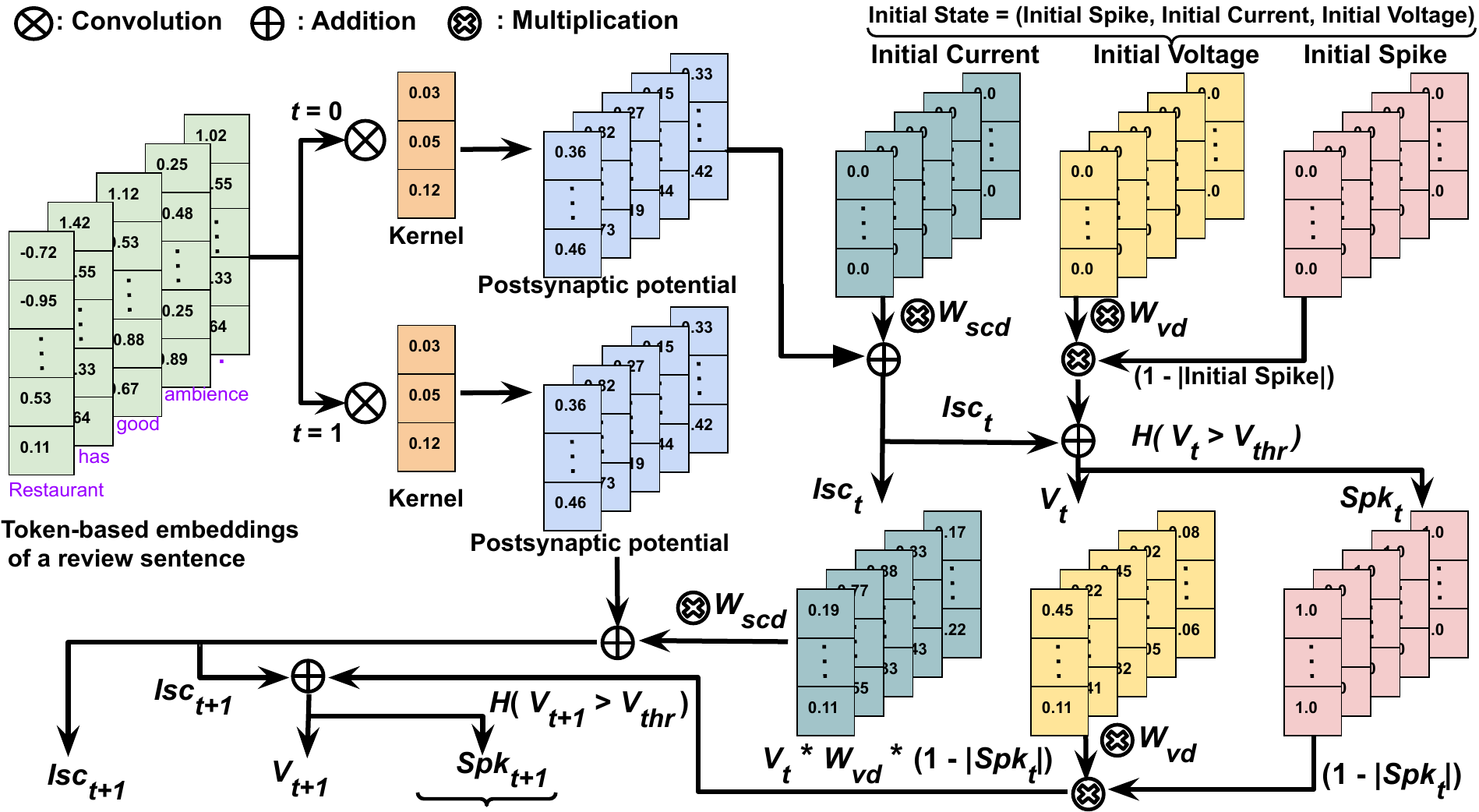}
\caption{Convolutional spike encoding operation for the first two time steps.} 
\label{fig:Conv_Spk_encd}
\vspace{-12pt}
\end{figure}

Figure~\ref{fig:Conv_Spk_encd} shows an example of this operation across two time steps. The tokenized input data \(S \in \mathbb{R}^{B \times R \times E}\), where \(B\) is the batch size, \(R\) is the sequence length, and \(E\) is the embedding dimension, are convolved with a set of filters to capture features along the sequence. The filters are represented by the weight tensor \(W^{enc} \in \mathbb{R}^{C \times E \times K}\), where \(C\) is the number of output channels and \(K\) is the kernel size, spanning both the embedding dimension and a fixed window of sequential tokens. The convolution operation computes each element of the output
tensor \(P \in \mathbb{R}^{B \times R \times C}\) by sliding the filters across the sequence length \(R\) in \(S\), performing a weighted sum of values in each window and adding a bias \(b \in \mathbb{R}^{C}\) for each channel. The result \(P\) a transformed representation of the input sequence that encodes local features and patterns across the sequence for each batch in \(B\) and each output channel in \(C\), with the same sequence length \(R\) based on padding (2) and stride (1) settings is calculated as
\begin{equation}\label{eq:Eq4}
P_{i, j, k} = \sum_{l=0}^{E - 1} \sum_{m=0}^{K - 1} S_{i, j+m, l} \cdot W^{enc}_{k, l, m} + b^{enc}_k,
\end{equation}
where \(i\) indexes the batch, \(j\) is the position in the sequence length \(R\), and \(k\) is the output channel.

The convolution operation generates the output \(P \in \mathbb{R}^{B \times R \times C}\), encoding the inputs as postsynaptic potential values. Each neuron within the resulting feature maps is then processed through the LIF operation cycle described by \eqref{eq:Eq1} and \eqref{eq:Eq2}. This converts continuous word embedding values into spike events ${P({Spk}_t)}_{i, j, k}$, while preserving spatial and temporal relationships within the data. Importantly, the encoding process is trainable, 
eliminating the need for manual parameter adjustment that is often required in conventional heuristic or rate-based encoding methods~\citep{schrauwen2003bsa, petro2019selection}.

\subsection{Spiking-Based Convolutional Layer} 
The dynamic spike patterns generated by the encoding are processed by this layer. It operates on spike inputs and uses synaptic weight adjustments in convolution operations to process them, capturing complex spatial and temporal dependencies and extracting higher-level features. This approach improves the interpretive capacity of the layer and also aligns closely with the dynamic and adaptive processing typical of neuronal systems. The convolution operation is given by 
\begin{equation}\label{eq:Eq5}
f_{conv}({P({Spk}_t)})_{i, j, k} =  \sum_{l=0}^{E - 1} \sum_{m=0}^{K - 1} {P({Spk}_t)}_{i, j+m, l} \cdot W^{L}_{k, l, m} + b^L_k,
\end{equation}
that calculates the postsynaptic potential for the feature maps in the $L^{\mathrm{th}}$ convolutional layer using learnable parameters (kernel weights $W^L$ and biases $b^L$) to fine-tune the network’s response at this layer. Building on this structure, our SNN architecture supports up to three additional spiking-based convolutional layers, which process the spike output from the initial spiking
convolutional layer. This design facilitates increasingly intricate spatial and temporal feature extraction across the SNN. \\
\noindent \emph{Non-spiking-based fully connected layer:} Similar to the output layer in a DNN, this layer serves as a decoder in the SNN. Its purpose is to transform the spike-based representation into a more interpretable output: a vector of class probabilities for each time step. 
{\small \begin{align}\label{eq:Eq7}
 & \underset{t}{logits} = W^L_{nspk} \cdot Spk^{L-1}_t + b^L_{nspk} \;,\;\; \mathrm{and} \;\; Prob_{class} = \sum_{t} Softmax(\underset{t}{logits}). 
\end{align}}
Here, $W^L_{nspk}$ and $b^L_{nspk}$ are the learnable parameters of the non-spiking fully connected layer, while $Spk^{L-1}_t$ represents the incoming spikes from the preceding spiking-based convolutional layer. Using these inputs, equation \eqref{eq:Eq7} calculates the time-specific vector of class logits, $\underset{t}{logits}$ for each token. The final class probabilities for each token,
$Prob_{class} \in \mathbb{R}^{B \times R \times |Y|}$, are then determined by aggregating across all time steps. In summary, the non-spiking fully connected layer consolidates and interprets the features learned by
previous layers, serving as a decoder that maps complex spike-based feature representations into comprehensible output predictions.

\subsection{Training Procedure for Spiking Neural Network}
The process of updating a neuron's state at time \( t \) is described in Algorithm~\ref{alg:LIF}. The Heaviside step function, \( H \), used in the algorithm, is defined as \( H(z) = 1 \) for \( z \ge 0 \) and \( H(z) = 0 \) otherwise. When a neuron fires a spike at time \( t-1 \), its membrane potential is reset at the next time step \( t \). This reset effectively removes the decayed component \( W_{vd}^L \cdot {V_{t-1}}^L \) using the term \( (1 - \big|{Spk}^L_{t-1}\big|)\), ensuring that the membrane potential is cleared in preparation for the next cycle of activity. The neuron's transition to the next state, denoted \( {State}^L_t \), follows a systematic cycle of updating, firing, and resetting, as governed by Equations \eqref{eq:Eq1} and \eqref{eq:Eq2}.
\setlength{\textfloatsep}{6pt}
\begin{algorithm} 
\setstretch{1.1}
\caption{LIF neuron state update at time $t$ in layer $L$}\label{alg:LIF}
\begin{algorithmic}[1]
\item[\textbf{Input:}] 
\item[${Spk}^L_{t-1}$: previous spike, ${Isc}^L_{t-1}$: synaptic current]
\item[$V^L_{t-1}$: membrane potential, ${Spk}^{L-1}_t$: current spike from]
\item[previous layer, $W_{scd}^L$: synaptic current-decay weight,]
\item[$W_{vd}^L$: voltage-decay weight, $V_{thr}$: threshold voltage ,]
\item[$W_{fv}$: postsynaptic potential parameters, $H$: heaviside]
\item[step function]
\item[] 
\Function{LIFStateUpdate}{${Spk}^{L-1}_t, {State}^L_{t-1}, $\par$ W_{scd}^L$, $W_{vd}^L, V_{thr}, W_{fv}, H$}
    \State ${Spk}^L_{t-1}, {Isc}^L_{t-1}, V^L_{t-1} \leftarrow {State}^L_{t-1}$
    \State ${Isc}^L_{t} \leftarrow W_{scd}^L \cdot {Isc}^L_{t-1} + W_{fv} \cdot {Spk}^L_{t-1}$
    \State ${V_t}^L \leftarrow W_{vd}^L \cdot {V_{t-1}}^L \cdot (1 - \big|{Spk}^L_{t-1}\big|) + {Isc}^L_t$
    \State ${Spk}^L_t \leftarrow H( {V_t}^L - V_{thr})$
    \State ${State}^L_t \leftarrow ({Spk}^L_t, {Isc}^L_{t}, {V_t}^L)$
    
    \State \Return ${Spk}^L_t$, ${State}^L_t$
    
\EndFunction
\end{algorithmic}
\end{algorithm}
SpikeATE is trained using a cross-entropy loss function, $\mathcal{L}_{ce}$, to classify three distinct token types $(B,I,O)$.
\begin{equation}\label{eq:LCE}
\small
    \mathcal{L}_{ce} = -\frac{1}{N} \sum_{i=1}^{N} \frac{1}{R_i} \sum_{j=1}^{R_i} \sum_{c=0}^{|Y|-1} \mathbf{1}(y_{i,j} = c) \cdot \log(\underset{i,j,c}{Prob_{class}})
\end{equation}

For each token $j$ in example $i$, $y_{i,j}$ denotes the true class label (from categories \{0, 1, 2\}), while $\underset{i,j,c}{Prob_{class}}$ represents the predicted probability that the token belongs to class $c$. The indicator function $\mathbf{1}(y_{i,j} = c)$ is 1 when the true label matches the class $c$ and 0 otherwise. This loss function averages the cross-entropy across all tokens and examples, providing a measure of how well the predicted probabilities align with the true labels in this classification task.

Since spike signals propagate across SNN layers in the spatial domain and also influence neuronal states over time, gradient-based training methods must consider derivatives in both spatial and temporal dimensions. We have integrated the synaptic current decay weights, $W_{scd}$, and the voltage decay weights $W_{vd}$ of the LIF neurons, along with other SNN parameters, into a spatio-temporal backpropagation method~\citep{wu2018spatio}. This approach minimizes the average cross-entropy loss, $\mathcal{L}_{ce}$, over all training samples, $N$.

The Heaviside step function $H$ which governs spike activation is not differentiable. Therefore, when calculating its gradient, we adapt the approach proposed by Fang et al.~\citep{fang2021incorporating} by an arctangent function $Atan(V) = \frac{1}{\pi} \arctan\left(\frac{\pi}{2} \alpha V\right) + \frac{1}{2}$ and using it to approximate the derivative of the spike activity with respect to membrane voltage $V$ as $G_{ps}(V) = \frac{\alpha}{2} \cdot \frac{1}{1 + \left(\frac{\pi}{2} \alpha V\right)^2}$, where $G_{ps}$ is the pseudo-gradient and $\alpha$ is a scaling factor that controls the steepness and spread of the arctangent function, allowing flexibility in shaping the function's responsiveness to $V$. Differentiation of the loss function $\mathcal{L}_{ce}$ with respect to the neuronal state variables is
{\small
\begin{align} 
\label{eq:gd_eqns}
    & \frac{\partial \mathcal{L}_{ce}}{\partial V^L_t} = G_{ps}(V^L_t) \cdot \small\frac{\partial \mathcal{L}_{ce}}{\partial {Spk}^L_t} + W^L_{vd} \cdot (1-\big|{Spk}^L_t\big|) \cdot \frac{\partial \mathcal{L}_{ce}}{\partial V^L_{t+1}},  \\
    & \frac{\partial \mathcal{L}_{ce}}{\partial {Isc}^L_t} = \frac{\partial \mathcal{L}_{ce}}{\partial {V}^L_{t+1}} + W^L_{scd} \cdot \frac{\partial \mathcal{L}_{ce}}{\partial {Isc}^L_{t+1}}, \frac{\partial \mathcal{L}_{ce}}{\partial {Spk}^{L-1}_t} = {W'}^L \cdot \frac{\partial \mathcal{L}_{ce}}{\partial {Isc}^L_t},\nonumber
\end{align}}
where $Isc$ is the synaptic current, $V$ is the voltage, and $Spk$ denotes the output spikes for each layer $L$. We also calculate the gradient of cross-entropy loss with respect to the SNN parameters for each spatial convolution and linear layer. This requires accumulating the backpropagated gradients over the entire sequence of timesteps 
\vspace{-4pt}
\begin{equation}\label{eq:grad_ts}
\small
    \frac{\partial \mathcal{L}_{ce}}{\partial {W}^L} = \sum_{t=1}^{T}Spk^{L-1}_t \cdot \frac{\partial \mathcal{L}_{ce}}{\partial {Isc}^L_t} \;,\;\; \mathrm{and} \;\; \frac{\partial \mathcal{L}_{ce}}{\partial {b}^L} = \sum_{t=1}^{T} \frac{\partial \mathcal{L}_{ce}}{\partial {Isc}^L_t}.  
\end{equation}
Finally, we calculate the derivatives of the loss function with respect to the synaptic current-decay weight $W_{scd}^L$ and the voltage-decay weight $W_{vd}^L$ which are the essential parameters within the SNN architecture for each layer $L$.
{\small
\begin{align}\label{eq:grad_decay}
    & \frac{\partial \mathcal{L}_{ce}}{\partial {W_{scd}}^L} = \sum_{t=1}^{T}{Isc}^L_{t-1} \cdot \frac{\partial \mathcal{L}_{ce}}{\partial {Isc}^L_t} \;,\;\; \mathrm{and} \;\;
    \frac{\partial \mathcal{L}_{ce}}{\partial {W^L_{vd}}} = \sum_{t=1}^{T} V^L_{t-1} \cdot (1-\big|{Spk}^L_{t-1}\big|) \cdot \frac{\partial \mathcal{L}_{ce}}{\partial {V}^L_t}  
\end{align}}

\begin{algorithm}[t!]
\setstretch{1.1}
\caption{Training Procedure}\label{alg:Training}
\begin{algorithmic}[1]
\item[\textbf{Input:}] 
\item[$T_{D}$: training set $\left\{ (S^i, Y^i) \right\}_{i=1}^{N}$ ]
\item[$W$: weights, $Spk$: spikes, $V_{thr}$: threshold voltage]
\item[$Isc$: synaptic current, $W_{scd}$: current-decay weight]
\item[$V$: membrane potential, $W_{vd}$: voltage decay weight]
\item[$\eta$: learning rate, $T$: timesteps, $n_{epochs}$: epochs]
\item[$W_{fv}$: postsynaptic function weight] 
\item[] 

\State $L^{Total}=$ [Convolutional Spike Encoding ($\cdot$), Spike-based Conv1 ($\cdot$), Spike-based Conv2 ($\cdot$), Spike-based Conv3 ($\cdot$))]
\State $L_{nspk}=$ Non-spiking-based fully connected ($\cdot$)
\For{each mini-batch B in $T_{D}$}
    \State \textbf{/* Forward pass */}
    \State Out\_prob = [ ] {/* List to store output probabilities */}
    \For{$t=1, \dots, T$}
        \For{$L$ in $L^{Total}$}
            \State /* Update ${State}^L_t$ = $({Spk}^L_t, {Isc}^L_{t}, {V_t}^L)$ of 
            \State neurons in layer $L$ at timestep $t$. */ 
            \State ${Spk}^L_t$, ${State}^L_t = $  LIFStateUpdate(${Spk}^{L-1}_t,$\par \hspace{45pt}${State}^L_{t-1}, W_{scd}^L$, $W_{vd}^L, V_{thr}, W_{fv}, H$)
           
        \EndFor
        \State {/* $L=L_{Total}[-1]$  */}
        \State  Out\_prob += $Softmax(L_{nspk}(Spk^L_t))$  
    \EndFor
    \State \textbf{/* Calculate the loss */}
    \State {/* Sum of output probabilities from all time steps */}
    \State Out\_prob = $\sum_{t=1}^{T} \text{Out\_prob}[t,:,:]$
    \State $\mathcal{L}_{ce}(y_B, \text{Out\_prob})$  {/* Out\_prob $\in$ (B, R, $|Y|$) */}

    \State \textbf{/* Backward pass */}
    \State Calculate gradient of the loss $\mathcal{L}_{ce}$ with respect to \par weights using \eqref{eq:gd_eqns}, \eqref{eq:grad_ts}, and \eqref{eq:grad_decay}.
    \State Update the model weights as $W := W - \eta \frac{\partial L_{ce}}{\partial W}$
\EndFor

\end{algorithmic}
\end{algorithm}

Algorithm~\ref{alg:Training} summarizes the training process. It comprises both forward and backward propagation phases across each epoch while iterating over each mini-batch of data. Line 1 creates a list $L^{Total}$ that contains all the instantiated layers within the network: the convolutional spike-encoding layer and multiple spike-based convolutional layers. Line 2 holds the instantiated non-spiking-based fully connected layer. Lines 3--22 describe how the training loop iterates through each mini-batch $B$ drawn from the training set $T_{D}$. The inner loop in lines 6--14 updates the state of LIF neurons within each layer for each time step $t$. The function $LIFStateUpdate()$, previously shown in Algorithm~\ref{alg:LIF}, is used to update each neuron's state within a layer. In line 13, spikes from the penultimate spike-based convolutional layer are transmitted to a non-spiking fully connected layer. Their outputs are integrated using a dot product to determine the time-specific vector of class logits, $\underset{t}{logits}$ for each token. To derive the final class probabilities, $Prob_{class}$ is calculated for each token at all time steps. Lines 17--18 calculate the loss using a cross-entropy function $\mathcal{L}_{ce}$, comparing the target output $y_B$ against the summation of probabilities over all time steps. Lines 20 and 21 describe the backward propagation phase: the loss gradient $\mathcal{L}_{ce}$ with respect to the weights is calculated according to \eqref{eq:gd_eqns}, \eqref{eq:grad_ts}, and \eqref{eq:grad_decay}. The model weights are then adjusted by subtracting a product of the learning rate $\eta$ and the calculated gradients.

To train the SpikeATE model, we used a batch size of 8 and a learning rate of 0.0001. The threshold voltage was set to 0.1. The model incorporates synaptic current and membrane voltage decay parameters, both set to 0.1, with a simulation time step of 6. Additionally, an arctangent scaling factor of 2 was applied to control the steepness and spread of the function, providing flexibility in its responsiveness to membrane potential. These settings collectively ensured stable training and efficient spike-based processing.




\section{Performance Evaluation}\label{sec:evaluation}
We compare the accuracy and computational efficiency achieved by SpikeATE with other competing approaches using datasets from SemEval 2014, 2015, and 2016~\citep{pontiki-etal-2014-semeval, pontiki-etal-2015-semeval,pontiki-etal-2016-semeval}. The reason for choosing the SemEval datasets is consistent with existing works~\citep{simmering2023large, varia2023instruction, wang2021progressive, mao2021joint, chen2020enhancing, li2020conditional, xu2019bert, ma2019exploring, xu2018double, li2018aspect, li2017deep, wang2017coupled, vicente2017elixa, toh2016nlangp, wang2016recursive, chernyshevich2014cross, toh2014dlirec}, which also employ these benchmarks for Aspect Term Extraction task. These datasets consist of review sentences in the restaurant and laptop domains with annotated aspect terms, each divided into separate training and test sets. To ensure a fair comparison with DNN-based approaches, we also held out 150 training samples as a validation set and for tuning purposes.

A key challenge in ATE on SemEval datasets is the limited availability of annotated data. To address this, previous research has employed various data augmentation  techniques, including conditional approaches, generating soft prototypes using internal and external data, labeling unlabeled data through self-training, and back-translation, among others~\citep{li2020conditional, chen2020enhancing, wang2021progressive, zhang2023target, ouyang2023unsupervised}. Our enhancement approach uses ChatGPT~\citep{assaraf2022openai} to predict aspect terms in the Amazon Cell Phones and Accessories dataset and the Yelp Review dataset~\citep{he2016ups, zhang2015character}, incorporating them as augmented data for the laptop and restaurant domains. We selected these datasets because they were also used for the augmentation process in Wang et al.~\citep{wang2021progressive} and share a similar context with laptop and restaurant reviews. After preprocessing, the summary of the datasets is shown in Table~\ref{table:ate_datasets}.

\begin{table}[t!]
\caption{Summary information for the datasets used in our evaluation.}
\label{table:ate_datasets}
\centering
\resizebox{\columnwidth}{!}{ 
\begin{tabular}{|c|cc|cc|cc|cc|}
\hline
\textbf{Datasets} & \multicolumn{2}{c|}{\textbf{Lap14}} & \multicolumn{2}{c|}{\textbf{Res14}} & \multicolumn{2}{c|}{\textbf{Res15}} & \multicolumn{2}{c|}{\textbf{Res16}}  \\ 
\hline
Type & Sentences & Aspects & Sentences & Aspects & Sentences & Aspects & Sentences & Aspects \\
\hline
Train & 3045 & 2342 & 3041 & 3686 & 1315 & 1209 & 2000 & 1757 \\
Test & 800 & 650 & 800 & 1134 & 685 & 547 & 676 & 622 \\
Train\_Aug & 3561 & - & 6186 & - & 6186 & - & 6186 & - \\
\hline
\end{tabular}
}
\end{table}

\subsection{Results and Ablation Study}
Table~\ref{table:performance_comparison} compares SpikeATE with other leading methods in terms of F1-score. Referring to the table, IHS-RD~\citep{chernyshevich2014cross}, DLIREC~\citep{toh2014dlirec}, EliXa~\citep{vicente2017elixa} and NLANGP~\citep{toh2016nlangp} are the benchmark winning models for the Lap14, Res14, Res15, and Res16 datasets, respectively~\citep{chernyshevich2014cross, toh2014dlirec,vicente2017elixa,toh2016nlangp}. Other approaches in this comparison employ diverse and sophisticated DNNs. The results for the first eight methods are sourced from Li et al.~\citep{li2020conditional}, while the other scores are averaged over three runs with random initialization. 

\begin{table}[t!]
\caption{Performance comparison on ATE datasets.}
\vspace{-5pt}
\label{table:performance_comparison}
\centering
  \small
\fontsize{10}{12}\selectfont
{
\begingroup
\setlength{\tabcolsep}{8.5pt} 
\renewcommand{\arraystretch}{.85} 
\begin{tabular}{lcccr}
\toprule
\textbf{Method} & \textbf{Lap14} & \textbf{Res14} & \textbf{Res15} & \textbf{Res16} \\
\midrule
IHS-RD~\citep{chernyshevich2014cross}  & 74.55 & 79.62 & - & - \\
DLIREC~\citep{toh2014dlirec} & 73.78 & 84.01 & - & - \\
EliXa~\citep{vicente2017elixa} & - & - & 70.04 & - \\
NLANGP~\citep{toh2016nlangp} & - & - & 67.12 & 72.34 \\
RNCRF~\citep{wang2016recursive} & 78.42 & 84.93 & 67.74 & 69.72 \\
MIN~\citep{li2017deep} & 77.58 & - & - & 73.44 \\
CMLA~\citep{wang2017coupled} & 77.80 & 85.29 & 70.73 & 72.77 \\
HAST~\citep{li2018aspect} & 79.52 & 85.61 & 71.46 & 73.61 \\
Seq2Seq4ATE~\citep{ma2019exploring} & 79.02 & 84.08 & 69.89 & 72.82 \\
DECNN~\citep{xu2018double} & 81.39 & 86.04 & 71.18 & 74.39 \\
CDA~\citep{li2020conditional} & 81.58 & - & - & 75.19 \\
SoftProtoE~\citep{chen2020enhancing} & 83.19 & 87.39 & 73.27 & 76.98 \\
BERT-RC~\citep{mao2021joint} & 81.84 & 85.27 & 70.16 & 75.47 \\
BERT-PT~\citep{xu2019bert} & 84.23 & 86.32 & 73.85 & 78.32 \\
Self-Training~\citep{wang2021progressive} & 86.91 & 88.75 & 75.82 & 82.56 \\
IT-MTL~\citep{varia2023instruction} & 76.93 & - & 74.03 & 79.41 \\
GPT-3.5~\citep{simmering2023large} & 83.8 & 83.8 & - & - \\
\midrule
\textbf{Ours (Binary)} & 81.35 & 84.85 & 69.98 & 75.33 \\
\textbf{Ours (Ternary)} & 84.02 & 86.46 & 72.25 & 78.19 \\
\bottomrule
\end{tabular}
\endgroup }
\end{table}

As seen in Table~\ref{table:performance_comparison}, the ternary variant of SpikeATE is highly competitive across all datasets, closely matching or exceeding the leading models in certain cases. 
Simmering et al. use GPT-3.5 in conjunction with a joint approach for aspect term extraction and polarity classification, incorporating both tasks during supervised training~\citep{simmering2023large}. They reported an F1 score of 83.8 in SemEval-2014 by combining the Lap14 and Res14 datasets. Consequently, we have recorded a score of 83.8 in the Lap14 and Res14 columns in Table~\ref{table:performance_comparison}. In contrast, our methodology focuses solely on aspect term extraction during training. Considering the broader category of large language models, SpikeATE outperformed GPT-3.5 and BERT-RC~\citep{mao2021joint}, exceeded BERT-PT~\citep{xu2019bert} on Res14, and achieved comparable F1 scores across other datasets. The above results indicate SpikeATE’s capability to handle domain-specific aspect extraction tasks with high accuracy. Overall, SpikeATE is an efficient alternative to DNNs, offering a robust balance between performance and model simplicity, and highlights the potential of SNNs for ATE tasks.

Table~\ref{table:ablation_study} presents an ablation study on the performance of binary and ternary spike-based models for four datasets, in terms of F1-score for different time steps (4 and 6). Ternary spikes improve model performance across all datasets and time steps due to the added expressive capability. For instance, on the Laptop14 dataset, the F1-score increases from 78.09 to 83.74 for 4 time-steps and from 81.35 to 84.02 for 6 time-steps when switching from binary to ternary spikes. Similarly, substantial improvements are observed in the Restaurant datasets, with the ternary model achieving higher F1-scores than the binary model for both time-step settings. 

\begin{table}[t!]
\caption{Performance of binary and ternary spikes.}
\label{table:ablation_study}
\centering
  \small
\fontsize{10}{12}\selectfont
  {
   \begingroup
\setlength{\tabcolsep}{8.5pt} 
\renewcommand{\arraystretch}{.85} 
\begin{tabular}{lccr}
\toprule
\textbf{Dataset} & \textbf{Method} & \textbf{Time-steps} & \textbf{F1-score} \\
\midrule
\multirow{4}{*}{Lap14} & Binary spike & 4 & 78.09 \\
 & Ternary spike & 4 & 83.74 \\
 & Binary spike & 6 & 81.35 \\
 & Ternary spike & 6 & 84.02 \\
\midrule
\multirow{4}{*}{Res14} & Binary spike & 4 & 83.80 \\
 & Ternary spike & 4 & 85.51 \\
 & Binary spike & 6 & 84.85 \\
 & Ternary spike & 6 & 86.46 \\
\midrule
\multirow{4}{*}{Res15} & Binary spike & 4 &  67.38\\
 & Ternary spike & 4 & 71.11 \\
 & Binary spike & 6 & 69.98 \\
 & Ternary spike & 6 & 72.25 \\
\midrule
\multirow{4}{*}{Res16} & Binary spike & 4 & 73.81 \\
 & Ternary spike & 4 & 76.57 \\
 & Binary spike & 6 & 75.33 \\
 & Ternary spike & 6 & 78.19 \\
 \bottomrule
\end{tabular}
\endgroup }
\end{table}
\begin{table}[t!]
\caption{Effect of varying convolutional layers on the performance of binary and ternary SNNs.}
\label{table:ablation_study1}
\centering
  \small
\fontsize{10}{12}\selectfont
  {
   \begingroup
\setlength{\tabcolsep}{8.5pt} 
\renewcommand{\arraystretch}{.85} 
\begin{tabular}{lccr}
\toprule
\textbf{Dataset} & \textbf{Method} & \textbf{\#Conv} & \textbf{F1-score} \\
\midrule
\multirow{8}{*}{Lap14} & Binary spike & 1 &  71.77\\
 & Ternary spike & 1 &  73.51\\
 \cmidrule{2-4}
 & Binary spike & 2 &  78.23\\
 & Ternary spike & 2 &  80.92\\
 \cmidrule{2-4}
 & Binary spike & 3 & 81.35 \\
 & Ternary spike & 3 &  84.02\\
 \cmidrule{2-4}
 & Binary spike & 4 &  80.88\\
 & Ternary spike & 4 &  83.23\\
 \midrule
 \multirow{8}{*}{Res14} & Binary spike & 1 &  75.55\\
 & Ternary spike & 1 &  78.17\\
 \cmidrule{2-4}
 & Binary spike & 2 &  81.33\\
 & Ternary spike & 2 &  83.08\\
 \cmidrule{2-4}
 & Binary spike & 3 & 84.85 \\
 & Ternary spike & 3 & 86.46 \\
 \cmidrule{2-4}
 & Binary spike & 4 &  83.97\\
 & Ternary spike & 4 &  85.95\\
 \bottomrule
\end{tabular}
\endgroup }
\end{table}

Table~\ref{table:ablation_study1} quantifies the impact of varying convolutional layers on the performance of SpikeATE with binary and ternary spikes across two datasets. Across all configurations, ternary spike models consistently outperform binary spike models in terms of F1-score, indicating that ternary representations enhance information capacity. For both datasets, increasing the number of convolutional layers improves performance, with the highest F1-scores observed up to three layers. Although performance deteriorates slightly with four layers, ternary spike models maintain an edge over binary models across all settings. This suggests that ternary SNNs benefit from deeper architectures but may reach diminishing returns beyond a certain depth, a phenomenon consistent with findings in the prior literature~\citep{rolnick2017power}.

\begin{table*}[t!]
\caption{Case study: Aspect terms are highlighted in purple.}
\label{tab:case_study_results}
\centering
  \small
\fontsize{10}{12}\selectfont
  {
   \begingroup
\setlength{\tabcolsep}{8.5pt} 
\renewcommand{\arraystretch}{.85} 
\begin{tabular}{rp{11cm}}
\toprule
\textnormal{\textbf{\textcolor{orange}{Review1}:}} & \textnormal{it is super fast and has outstanding \textcolor{custompurple}{graphics}.} \\
\hline
\textnormal{Tokens:} & \textnormal{['it', 'is', 'super', 'fast', 'and', 'has', 'outstanding', '\textcolor{custompurple}{graphics}', '.']} \\
\hline
\textnormal{BIO:} & \textnormal{['O', 'O', \hspace{5pt}'O', \hspace{7pt} 'O', \hspace{5pt}'O', \hspace{5pt} 'O', \hspace{18pt}'O', \hspace{35pt}'\textcolor{custompurple}{B}', \hspace{5pt}'O']} \\
\hline
\textnormal{$\widehat{\textnormal{BIO}}$ (Pred):} & \textnormal{['O', 'O', \hspace{5pt}'O', \hspace{7pt} 'O', \hspace{5pt}'O', \hspace{5pt} 'O', \hspace{18pt}'O', \hspace{35pt}'\textcolor{custompurple}{B}', \hspace{5pt}'O']}  \textcolor{green}{\cmark}   \\

\midrule

\textnormal{\textbf{\textcolor{orange}{Review2}:}} & \textnormal{the \textcolor{custompurple}{mountain lion os} is not hard to figure out if you are familiar with \textcolor{custompurple}{microsoft windows}.} \\
\hline
\textnormal{Tokens:} & \textnormal{['the', '\textcolor{custompurple}{mountain}', '\textcolor{custompurple}{lion}', '\textcolor{custompurple}{os}', 'is', 'not', 'hard', 'to', 'figure', 'out', 'if', 'you', 'are', 'familiar', 'with', '\textcolor{custompurple}{microsoft}', '\textcolor{custompurple}{windows}', '.']} \\
\hline
\textnormal{BIO:} & \textnormal{['O', \hspace{17pt}'\textcolor{custompurple}{B}', \hspace{21pt}'\textcolor{custompurple}{I}', \hspace{4pt} '\textcolor{custompurple}{I}', \hspace{3pt}'O', \hspace{2pt}'O', \hspace{7pt}'O', \hspace{5pt}'O', \hspace{7pt}'O', \hspace{9pt}'O', \hspace{2pt}'O', \hspace{3pt}'O', \hspace{5pt}'O', \hspace{15pt}'O', \newline \hspace{18pt}'O', \hspace{21pt}'\textcolor{custompurple}{B}', \hspace{35pt}'\textcolor{custompurple}{I}', \hspace{7pt}'O']} \\

\hline
\textnormal{$\widehat{\textnormal{BIO}}$ (Pred):} & \textnormal{['O', \hspace{17pt}'\textcolor{custompurple}{B}', \hspace{21pt}'\textcolor{custompurple}{I}', \hspace{4pt} '\textcolor{custompurple}{I}', \hspace{3pt}'O', \hspace{2pt}'O', \hspace{7pt}'O', \hspace{5pt}'O', \hspace{7pt}'O', \hspace{9pt}'O', \hspace{2pt}'O', \hspace{3pt}'O', \hspace{5pt}'O', \hspace{15pt}'O', \newline \hspace{18pt}'O', \hspace{21pt}'\textcolor{custompurple}{B}', \hspace{35pt}'\textcolor{custompurple}{I}', \hspace{7pt}'O']} \textcolor{green}{\cmark}\\

\midrule

\textnormal{\textbf{\textcolor{orange}{Review3}:}} & \textnormal{i had the \textcolor{custompurple}{thai style fried sea bass} ... which was very good.} \\
\hline
\textnormal{Tokens:} & \textnormal{['i', 'had', 'the', '\textcolor{custompurple}{thai}', '\textcolor{custompurple}{style}', '\textcolor{custompurple}{fried}', '\textcolor{custompurple}{sea}', '\textcolor{custompurple}{bass}', '...', 'which', 'was', 'very', 'good', '.']} \\
\hline
\textnormal{BIO:} & \textnormal{['O', 'O', \hspace{3pt} 'O', \hspace{4pt}'\textcolor{custompurple}{B}', \hspace{12pt}'\textcolor{custompurple}{I}', \hspace{13pt}'\textcolor{custompurple}{I}', \hspace{13pt}'\textcolor{custompurple}{I}', \hspace{12pt}'\textcolor{custompurple}{I}', \hspace{5pt}'O', \hspace{6pt}'O', \hspace{13pt}'O', \hspace{7pt}'O', \hspace{11pt}'O', \hspace{1pt}'O']} \\

\hline
\textnormal{$\widehat{\textnormal{BIO}}$ (Pred):} & \textnormal{['O', 'O', \hspace{3pt} 'O', \hspace{4pt}'\ul{O}', \hspace{11pt}'\ul{O}', \hspace{10pt}'\textcolor{custompurple}{\ul{B}}', \hspace{10pt}'\textcolor{custompurple}{I}', \hspace{12pt}'\textcolor{custompurple}{I}', \hspace{5pt}'O', \hspace{6pt}'O', \hspace{13pt}'O', \hspace{7pt}'O', \hspace{11pt}'O', \hspace{1pt}'O']} \textcolor{red}{\xmark}\\
\bottomrule
\end{tabular}
\endgroup }
\end{table*}
\subsection{Detailed Analysis}
We more closely evaluated SpikeATE's effectiveness by analyzing examples from the test set (Table~\ref{tab:case_study_results}). The model demonstrates a strong tendency for aspect extraction, successfully identifying single-word aspects (e.g., ``graphics'' in Review1), two-word aspects (e.g., ``microsoft windows'' in Review2), and three-word aspects (e.g., ``mountain lion os'' in Review2). However, it struggles with longer aspect phrases, as seen in Review3, where ``thai style fried sea bass'' was only partially identified, with ``thai'' and ``style'' misclassified as outside (O). This suggests that while the model effectively captures contextual dependencies, it requires further refinement in recognizing phrase boundaries and aspect initiation. In Review3, SpikeATE successfully identified the latter part of the aspect phrase (``fried sea bass'') but failed to recognize the first two
tokens (``thai'' and ``style''). This implies that the spike train responsible for detecting aspect boundaries was either too weak or
insufficient. A possible reason for this misclassification is the imbalance in the training data, where only about 5\% of aspect
phrases contain four or more tokens. This disparity may have caused the model to bias its synaptic weights toward shorter aspect terms,
making them more easily detectable.

Ideally, when a multi-word aspect phrase such as ``thai style fried sea bass'' appears in a sentence, the first token (B-token) should generate a strong enough spike response to ensure the correct classification of itself and subsequent tokens (I-tokens). However, the model's predictions suggest that the spike activity for ``thai'' and ``style'' did not reach the required threshold, preventing them from being classified as aspect-term parts. Instead, the third token (``fried'') was classified as a B-token, likely because it received a higher spike response than the preceding tokens. 

SpikeATE is a brain-inspired design in which underlying spiking neurons mimic biological neuronal behavior. As observed in Review3,
maintaining short-term dependencies is easier in both artificial and biological neurons, often leading to recency bias~\citep{escobar2007long}. However, long-term dependencies pose greater challenges due to the complex interplay of synaptic plasticity, protein turnover, and the need for repeated reinforcement in biological systems~\citep{abraham2008metaplasticity}. Drawing from this neurological analogy, we could mitigate model bias by reinforcing multi-word aspect terms more effectively. One approach in future work would be to introduce augmentation techniques that ensure a more balanced distribution of multi-aspect terms across training batches, thereby strengthening the model's ability to recognize longer dependencies consistently.

\begin{figure}[t!]
\centering
\includegraphics[width=0.75\textwidth]{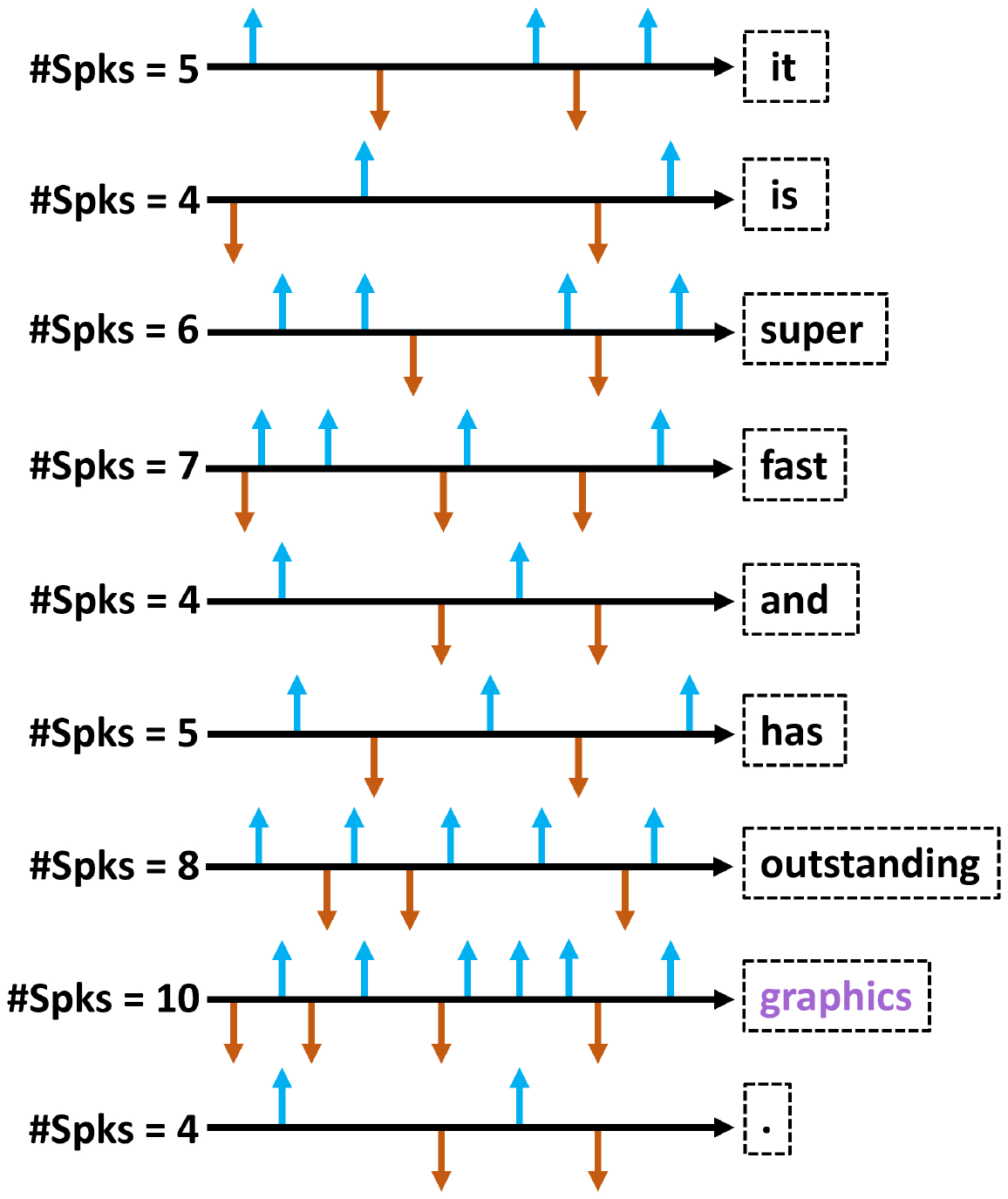}
\vspace{-35pt}
\caption{Visualization of spiking activity of the SNN when presented with tokens from Review1.} 
\label{fig:Spike_visualization}
\end{figure}

Figure~\ref{fig:Spike_visualization} shows the network's spiking activity when presented with tokens from Review1: ``it is super fast and has outstanding graphics,'' where incoming spikes from the previous convolutional layer are processed before being passed to a non-spiking fully connected layer. The upward blue arrows represent positive spikes, while the downward brown arrows indicate negative spikes. The aspect term ``graphics'' receives the highest number of spikes (10), highlighting its importance in the model's decision-making process. This suggests that SpikeATE has identified ``graphics'' as the most relevant term. The dominance of spikes for token ``graphics'' indicates that SpikeATE effectively focuses on aspect terms, helping to extract meaningful semantic features. 


\subsection{Computational Efficiency}
Neuromorphic hardware platforms promise ultra-low-power, event-driven computation. Prototypes have been developed and many other designs are under active development---for example, DYNAP-SE~\citep{Moradi_etal18}, Loihi~\citep{davies2018loihi}, Akida~\citep{akida2}, and TrueNorth~\citep{akopyan2015truenorth}, as well as various FPGA-based architectures~\citep{abdelsalam2018efficient,corradi2021,li2021fast,liu2023low,carpegna2022spiker,matinizadeh2024fully}. However, the commercial availability and accessibility of these chips is still limited. Consequently, recent work on developing SNNs for various applications (e.g., vision, control, text) use power models to estimate the computational efficiency of SNNs versus the traditional architectures they are intended to replace \citep{zhang2025staa, shibo2025ts, zhaottfsformer, guo2024ternary}. We use the same power model in our work to estimate efficiency as well.

The energy consumed during inference is modeled in terms of floating point operations (FLOPs) and synaptic operations (SOPs). The encoding and final layer use FLOPs, whereas other layers are evaluated by SOPs. For an SNN, the energy for a layer $L$ is calculated as $\text{Power}(L) = 77 \text{fJ} \times \text{SOPs}(L)$, where 77 fJ is the energy per SOP, and one sign needs 3.7 \text{pJ} based on empirical data~\citep{indiveri2015neuromorphic, hu2021spiking, guo2024ternary}. In binary SNN, only +1 or 0 are considered spikes. However, in ours, both +1 and -1 are treated as spikes, leading to a 
higher spike activity and consequently to more sign operations.
The number of SOPs in layer $L$ is $T \times \gamma \times \text{FLOPs}(L)$, with $T$ as simulation time steps, $\gamma$ as the mean spike firing rate in layer $L$, and $\text{FLOPs}(L)$ as the floating point operations for that layer. The firing rate $\gamma$ is $\gamma = \frac{N_{Spk}}{T}$, where $N_{Spk}$ is the spike count over $T$. For DNNs, the energy for layer $L$ is estimated as $\text{Power}(L) = 12.5 \, \text{pJ} \times \text{FLOPs}(L)$. FLOPs within the network are calculated as a sum of 
\begin{align} \label{eq:flops}
    & {FLOPs}_{conv} = c^L \times d^L \times w_{c^L} \times h_{c^L} \times w_{w^L} \times h_{w^L} \times 2, \nonumber \\
    & {FLOPs}_{fc} =  u^L \times u^{L-1} \times 2,
\end{align}
where ${FLOPs}_{conv}$ denotes the number of floating point operations for convolutional layers and ${FLOPs}_{fc}$, the number
of calculations for fully connected layers. The following parameters are used to calculate the computational
requirements of each layer within the network: $c^L$, the number of output feature maps within layer $L$; $d^L$, the number
of input channels; $w_{c^L}$ and $h_{c^L}$, the width and height of the output feature maps, respectively; $w_{w^L}$ and
$h_{w^L}$, the width and height of the convolutional kernels, respectively; and $u^L$, the number of neuron units present
in layer $L$. This formulation provides a comprehensive framework for estimating the energy consumption and computational
requirements of SNNs and DNNs, enabling a direct comparison of their efficiency.
\begin{table}[t!]
\caption{Theoretical estimation of energy consumption.}
\label{tab:th_energy_results}
\centering 
 \small
\fontsize{10}{12}\selectfont
  {
   \begingroup
\setlength{\tabcolsep}{8.5pt} 
\renewcommand{\arraystretch}{.85} 
\begin{tabular}{lcr}
\toprule
\textbf{Method} & \textbf{FLOPs / SOPs ($10^9$)} & \textbf{Power (mJ)} \\
\midrule
HAST & 0.5232 & 6.5412\\
Seq2Seq4ATE & 2.4888 & 31.1104 \\
DECNN & 0.2580 & 3.2256 \\
CDA & 8.5409 & 106.7618 \\
SoftProtoE & 0.2580 & 3.2256 \\
BERT-RC & 7.6448 & 95.5599  \\
BERT-PT & 7.6451 & 95.5636  \\
Self-Training & 7.6451 & 95.5636\\
GPT-3.5 & 3.14 x $10^{14}$ & 3.92 x $10^{15}$\\
\midrule
Ours (Binary) & 0.1152 / 0.0059 & 1.8943 \\
Ours (Ternary) & 0.1152 / 0.0149 & 2.5946  \\
\bottomrule
\end{tabular}
\endgroup }
\end{table}

Table~\ref{tab:th_energy_results} compares theoretical energy consumption across various models, highlighting GPT-3.5 as the most computationally
expensive with $(3.14 \times 10^{14})$ FLOPs and $(3.92 \times 10^{15})$ mJ power usage. BERT-based models and
Self-Training consume ~95.56 mJ, while HAST and DECNN are more efficient with power usage below 10 mJ.
The proposed binary SNN model achieves a power consumption of 1.8943 mJ, while the ternary model slightly increases to
2.5946 mJ, remaining highly efficient. The results indicate that spiking-based models significantly reduce energy
consumption compared to transformer-based architectures. This reinforces the advantage of spiking representations
for low-power NLP applications. 

\subsection{Latency and Throughput}
Latency is defined as the time interval between when a review sentence is presented to the SNN and when it is classified. SNNs process information based on spikes, which inherently introduces a form of temporal dependency not seen in typical DNNs. Given an input sentence, the SNN must wait for the spiking activity within neurons to reach steady state before proceeding with computation or making a decision, potentially increasing overall latency. When tested on a system with 2.00GHz Intel Xeon CPU and an NVIDIA Tesla T4 GPU, the inference time is approximately 1.4 ms. Despite the increased latency for predicting a review sentence, the low power consumption profile of SpikeATE allows for parallelization. 
Given a batch of review sentences, multiple SpikeATE instances can operate in parallel to predict these reviews to increase throughput. Hardware accelerators that efficiently process spikes, such as the Akida processor~\citep{akida1},  can be used to deploy SpikeATE. 

\section{Conclusion}\label{sec:conclusion}
We have developed SpikeATE, an SNN for ATE that advances NLP in an energy-efficient direction. Using a sparse spiking mechanism, ternary spiking neurons, and a custom convolutional spike encoding layer, SpikeATE effectively captures spatial and temporal dependencies while reducing computational demands.  Our backpropagation method is designed for spike-based networks, improving training within the SNN framework. Experimental results on four SemEval benchmark datasets confirm that SpikeATE achieves high precision in aspect term recognition, performing comparable to state-of-the-art DNNs but with significantly lower energy consumption.  

Future work will aim to unleash the full potential of SpikeATE in terms of energy efficiency and real-time performance using neuromorphic chips, either off-the-shelf (e.g., Akida) or FPGA-based designs~\citep{matinizadeh2024neuromorphic}.

\bibliographystyle{ACM-Reference-Format}
\bibliography{AspectTerm}










\end{document}